\documentclass{ws-ijig}

\begin{document}

\markboth{Pierre-Alexandre FAVIER and Pierre DE LOOR}
{From decision to action~:~intentionality, a guide for the specification of intelligent agents' behaviour}

\catchline{}{}{}{}{}
%
%%%%%%%%%%%%%%%%%%%%%%%%%%%%%%%%%%%%%%%%%%%%%%%%%%%%%%%%%%%%%%%%%%%%

%\title{Intentionality~:~a needed link between decision and action for intelligent agents' behaviour}
\title{From decision to action~:~intentionality, a guide for the specification of intelligent agents' behaviour}

\author{Pierre-Alexandre FAVIER}
\author{Pierre DE LOOR}
\address{Laboratoire d'Informatique des SYstèmes Complexes \\ Centre Europ\'een de R\'ealit\'e Virtuelle\\ 25 rue Claude Chappe \\ 29280 Plouzan\'e (France)\\ \{favier,deloor\}@enib.fr}

\maketitle
\setcounter{page}{87}
%\begin{history}
%\received{(Day Month Year)}
%\revised{(Day Month Year)}
%\accepted{(Day Month Year)}
%\comby{(xxxxxxxxxx)}
%\end{history}

\begin{abstract}
This article introduces a reflexion about behavioural specification for interactive and participative agent--based simulation in virtual reality. Within this context, it is necessary to reach a high level of expressivness in order to enforce interactions between the designer and the behavioural model during the in--line prototyping. This requires to consider the need of semantic very early in the design process. The Intentional agent model is here exposed as a possible answer. It relies on a mixed imperative and declarative approach which focuses on the link between decision and action. The design of a tool able to simulate virtual environment implying agents based on this model is discussed.\\
\end{abstract}

\keywords{multi--agent systems; intentional agent; in--line prototyping; behavioural specification; cross--paradigm programming.}

\section{Introduction}
This article presents a work in progress about behavioural specification for multi--agent systems and the study of a tool designed to interact with such systems. This is a part of the research area of the LI2\footnote{Laboratoire d'Ing\'enierie Informatique (Computer science Engineering Laboratory)} which is concerned by the study of methodologies and tools for agents' behaviour specification and execution, within the context of interactive and participative simulation. Our aim is to enable in--line prototyping of such agents~:~we want the end--user to be able to test and modify his model (agent's behaviour) at simulation time. 
This work introduces a reflexion around our agent--oriented language
\protect{\tt{oRis}}.\cite{1} This is an interpreted language wich is able to schedule the different activities of many agents and to dynamically modify the code of each instance through a run--time code parsing mechanism. \protect{\tt{oRis}}, combined with a dedicated rendering library, was succesfully used for really heterogenous applications dealing with domains as different as biology,\cite{2,3}dynamic problem solving systems, \cite{4}interactive simulator \cite{5} or virtual
environment for training.\cite{6,7} 
This last application field, virtual environment for training, is typically the kind of application we are targeting at~:~they imply need of a very high level specification since they need design of avatar with complex behaviours. If the end user is a teacher of a really specific domain, we can't expect him to be a multi--agent systems' expert at the same time. The design tool should provide an intuitive interface, allowing the designer to remain as close as possible to his domain ontology during the whole process.

This implies that the model should accept domain specific keywords as symbols for the behaviour specification. The user should therefore be able to specify and modify his model, manipulating his everyday vocabulary. This is not only a matter of interface, it implies the usage a behavioural model which can manipulate domain dependent symbols during the specification process on one hand, and on the other hand which can produce a ``feedback'' to the user with those symbols too~:~if the end user isn't a computer science expert, the system have to be able to produce on the fly some explanations using his vocabulary. This need of a semantic regards both the design process of the behaviour \protect{\bf{and}} its execution. This consideration must be kept in mind in order to propose a kind of behavioural model which favors high level interactions with the end--user and the associated execution tool.  

\protect{\tt{oRis}} was a partial answer to this problem~:~it allows on--line prototyping, but it requires the end user to be familiar with its syntax, wich is very close to \protect{\tt{C++}}. The reflexion exposed here deals with a new behavioural specification model proposition, and with the prerequities of the tool capable to execute it~:~we rely on the experience aquired through the building and the usage of \protect{\tt{oRis}} to create a new framework for the design and the execution of multi--agent systems closer to the end user.

As it was previously said, the building of such a tool can't be separated from the elaboration of the behavioural model~:~these two aspects of this approach emulate each other. 

Next sections of this article details our approach as a compromise~:~our desired mixing of reactive and symbolic approaches of behaviour specification induces a cross--paradigm (imperative and declarative) implementation of the execution model. Intentionality is introduced as a link between those different aspects of the behaviour specification and the intentional agent model is described. Finally, the first implementation of the framework is explained and some of the potential evolutions of this proposition are discussed as a conclusion.

\section{Need of a Specification Framework}
\subsection{From conception to execution}
Among the numerous methodologies that implement an agent's behaviour in a virtual reality simulation\cite{8}, it's usually very hard to chose the most appropriated one. Actually, this choice is rather a matter of habits of the designer or of industrial constraint. But the designer is not always the end--user~:~the implementation of a behaviour is a ``translation'' from the targeted domain to a definition in a computer language. Sometimes the computer science expert is able to learn enougth about the application domain to process coherently, but most of the time he has to interpret the model expressed by an expert of the application domain. Another approach is to ask this expert to learn computer science and to perform to the implementation. Those ways are not suitable for complex systems~:~they imply very specific knowledge from both the programmer and the designer. Though, a tool built for the design of such systems has to provide an interface between those two aspects of the implementation process usable by all the involved person. This link between the software implementation and the conceptual design of the behaviour should be used on one hand at design time, in order to help the communication within the developpment team, and on the other hand at run--time if we want to enable on--line prototyping~:~if we expect the designer (the end user) to interactively modify his model, we should provide a framework which can support the execution of agents specified and manipulated through domain dependent concepts; we can't ask an ethologist who is tuning a simulation involving fishes to dynamically modify the \protect{\tt{java}} class implementing their behaviour. 

This in--line prototyping objective implies a coherence from specification to execution~:~the model must be defined and manipulated (executed) the same way; definition and execution are not different steps anymore. So, our aim is to reach this complete coherence in the whole specification process in our framework and, ideally, to build a complete methodology of behaviour specification based on this framework.\\
A framework capable of such an abstract control level should rely on very generic mechanisms. The definition of the necessary set of those mechanisms is the main obstacle, there is no unique correct choice. Since the priority is given to the possibility of a manipulation through the domain dependent concepts and vocabulary, the class of expressible behaviour has to be restricted in order to define a minimal and consistent set of basics behavioural mechanisms to be implemented in the framework. The following describes the kind of behavioural model we expect to specify and execute.

\subsection{Meaning of action}
Our approach of behaviour specification is a compromise between reactive approaches\cite{9,10,11,12,13} and symbolical approaches\cite{14,15,16,17}.\\
The former are usually simple and easy to process automatically, and they are often a efficient for real--time rendering which is an important aspect in interactive simulation of virtual environment. Nevertheless they usually are not suitable for in--line prototyping because of lack of semantic in the model. The later, symbolical approaches, bring expressiveness and a good abstraction level; they are more confortable for non computer--science expert though they suffer from completeness and synchronization problem within the context of dynamic and open systems such as virtual environment~:~execution time for complex symbolic resolution could lead to coherence problem. 

To get rid of as many drawbacks of those existing approaches as possible, the one considered here is a mix of reactive and symbolic approaches~:~it's an action selection method which adds some semantical information about \protect{\it{meaning}} of those actions.\\
We don't want to provide a logic about action as in \cite{18,19,20} but we constraint the designer to add to every implemented method the needed symbols for reasoning. Each action (including perceptions which are here considered as particular actions) is in charge of the provision of some information about its side effect on the agent --- some of its properties --- in a qualitative way. This enables a symbolical reasoning about effects of actions which are defined following a reactive approach. \\

This first simple definition of our approach might look ambiguous, but it is a sufficient basis to discuss some important aspects of its implementation before its refinment, which is the next section purpose.

\subsection{A cross--paradigm implementation}
If we consider the behaviour specification as a translation from the application domain into a language computable by the machine, as previously suggested, we have to choose very carefully the final implementation language in order to keep a strong coherence within those different aspects. 

Some computer languages are very close to the machine model. Those are imperative languages which need the user to be familiar with very technical concepts such as memory allocation, assignement,\ldots They are difficult to use correctly but allow very efficient processing because they are really close to the hardware structure of the computer. They might be a good choice for 3D rendering and immersion in a virtual environment, but they are not suitable for an easy specification of complex behaviour~:~the building of an application enabling of in--line prototyping from scratch is a matter of specialist,\cite{21} not of the domain expert. It is really hard to design an agent capable of dynamic modification of its own code.\\ 

Some other languages were designed to be close to the end--user and to favour a high level abstraction, trying to reach the natural language as interface between machine and human beeing. Those are for example declarative or functional languages, such as \protect{\tt{Prolog}}. They are closer to our natural reasoning than \protect{\tt{C++}} or \protect{\tt{Java}}. They bring the expressiveness we expect for the end--user. They are not nevertheless suitable for the definition of side effects of the agent's action on itself and its environment in a dynamic context because of their performances. But the specification of a behaviour as an enumeration of behavioural rules seems really interesting~:~with a language such as \protect{\tt{Prolog}} the knowledge base is totally independent from the inference mechanism. One can deeply modify a logic program adding or deleting some clauses, even at run time. This is typically the kind of flexibility which can enable a high level in--line prototyping.\\

We have chosen to use both imperative and declaratice programming~:~imperative programming for the definition of side effects and perception (methods' implementation) together with declarative logic programming for the reasoning (action selection).\\

\begin{figure}[!th]
\begin{center}
\centerline{\psfig{file=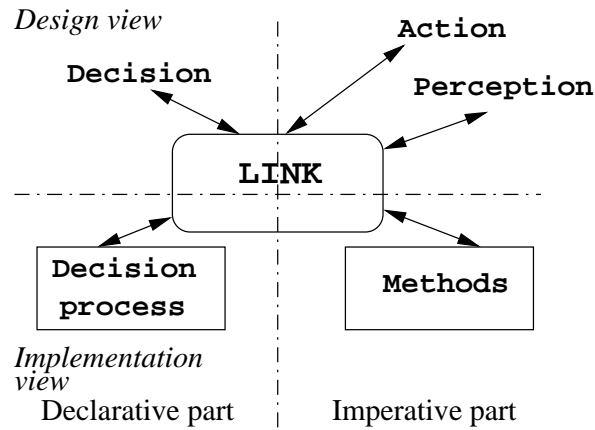}}
\vspace*{8pt}
\caption[fig1]{\label{fig1}Suggested model architecture}
\end{center}
\end{figure}

In order to keep the usage of such a tool simple, we have to provide a link between these two parts of the behaviour specification. For the framework's coherence, and to maximize expressiveness in the model, this link must be an automatic and implicit mechanism providing an interface between imperative and declarative parts. Moreover it must also be a therorical link between those two aspects of the specification~:~it must link decision part with action and perception part from the designer point of view, and, relying on the same concept, it must link the imperative and the declarative parts from the implementation point of view. The figure \ref{fig1} illustrates the role of this link, which is the main point of our approach.\\

The next section introduces the concept of intentionality as a possible symbolical link between the different parts of this framework.

\section{Intentionality as a Semantical Link}

\subsection{A mixed imperative / declarative programming approach}
The choices exposed in the previous sections are the fundation of the intentional model. In one sentence we could say that it's an action selection based on the qualitative explicitation in each action of its effects on the agent itself and its environment.\cite{22} Actions and perceptions are coded following an imperative approach (methods) whereas decision (action selection) is made by an inference engine which exploits a knowledge base filled in with behavioural rules. To keep coherence of the agent's state, the decision part must not modify the agent or its environment~:~only actions are able of side effect. This way, the logic part implied in the decision process is kept as ``pure'' as possible (no extra--logic predicates needed for some side effects), and a separation is deeply layed down between the action specification, which is the computer science expert job, and the behaviour specification, which is the role of the designer (our end--user). The decision process only needs to consult the value of the agent's properties and must select an action, or a set of actions, which best fit the situation. This is the unique interface between the two parts which has to be defined using a symbolic aproach. This favours the setting of a very meaningfull semantic link between those two parts from specification to execution~:~the needed interface between imperative and declarative part is entirely defined and implemented in the framework; the designer is constrained to follow our approach if he wants to use this tool. This is the first step through the building of a behaviour specification methodology.\\

\begin{figure}[!th]
\begin{center}
\centerline{\psfig{file=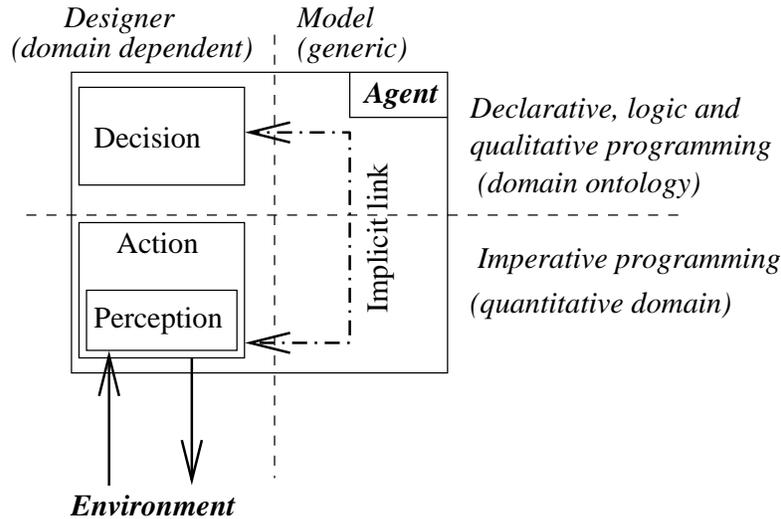}}
\vspace*{8pt}
\caption[fig2]{\label{fig2}Suggested model architecture}
\end{center}
\end{figure}

Figure \ref{fig2} illustrates the behavioural model architecture as it has been described up to this point of this paper. We now have to define the implicit link between the two parts of the behaviour specification.\\

The implicit link represented in this figure must not be specified by the user. It has to be entirely supported by the plateform. This ensures greater independence between the two aspects of the behaviour specification, which is required by the need of expressiveness for in--line prototyping.\\ 
Of course, the decision process can't be separated from the agent's state and environment or it wouldn't be accurate. Therefore, although the decision process can't modify the agent's properties, it must still be able to consult them at any time~:~one of the main feature of the framework should be the ability to consult on the fly the value of a property from the reasoning process. Initializing a logic inference with properties' values and wait for the result is not a good idea~:~as it was previously stated, logic inference process could be very slow regarding to the global simulation. Thereby, the decision part of the behaviour should be processed in parallel to agent's methods (actions and perceptions), and should be able to consult the value of a property at the exact time it's needed~:~during the whole process, which may implies many backtracks, the value of a property could become obsolete.\\
Perceptive actions are in charge of updating of those properties' values. This is the place where to put some domain dependent semantic in the model in order to increase expressiveness for the end--user. As every perception method is specific to an agent, its implementation can support the symbolic conversion of a property's value accordingly to domain vocabulary, \protect{\it e.g.~}:~whatever is the unit system used in the simulation plateform, the symbol \protect{\it{nearPredator}} can be updated  in the method \protect{\it{lookAround()}} of the agent, if this symbol is meaningfull for this kind of agent.\\
This requires from the user the enumeration of all the needed symbols for the decision process at design time. As those symbols depend on agent's state, they can be exhaustively enumerated~:~whatever your agent can see in a dynamic environment, the corresponding symbols (\protect{\it{prey, predator, obstacle,\ldots}}) are statically designed. This avoids the completeness problem of pure symbolical approaches in dynamic environment, and brings some more semantic in pure reactive approaches. Of course, the inference capabilities is limited by this static enumeration\\

To enhance coherence of the whole specification process, this link must not only be a link between decision part and action part of the model, but, what's more, a semantical link between specification and execution of the behaviour. So, if a symbolic exchange between decision and action is the technical selected choice, we now have to chose the semantical meaning of such a link. This choice will define the range of model which can be expressed and execute by this framework. Next section introduces the notion of intentionality as a first proposition for this semantical choice.

\subsection{Intentionality}
It has been previously explained that, in this approach, methods  are in charge of the production of symbols needed by the decision process. In order to obtain a systematic execution process of behaviour defined this way, and to progress in the building of a specification methodology laying on this framework, the semantic of those symbols has to be exactly defined.\\
The chosen class of expressible behaviour could be qualified as ``\protect{\it{rational behaviour}}''~:~we won't focus on pure reflex behaviour, neither on behaviour based on complex reflexion process. In our cross topic of virtual environments and agent--based simulations, we want to obtain adaptable and credible behaviour. This means that an agent is expected to select the ``good'' method, accordingly to its dynamic context. But this selection of the appropriated method can't be processed without explicitly expressing the \protect{\it{meaning}} of every method in the behavioural model itself. A behaviour definition is frequently very alike an enumeration of use case, selecting directly a method from some attributes using a \protect{\it{if\ldots then}}--style structure. This implies that the meaning of each method is only known at design time and totally disapears at simulation time~:~action selection is hard coded. This prevents the usage of this meaning within the agent, which yet would be part of in--line prototyping, or to generate some explaination automatically\ldots . Our approach is to introduce a symbolic expression of the \protect{\it{meaning}} of every action as an indirection between decision and action~:~the designer must provide all the information needed by a \protect{\bf{generic}} decision process to select an action. This obligation is a first step for the building of a design methodology. The set of behaviour expressible in this framework is limited by the generic aspect of this decision process.\\

The selection process must be template and does not make use of any logic of action~:~selection is only based on the context and known \protect{\it{meaning}} of actions. But the context is perceived through perception methods, so it is totally defined by the way of agent's properties (internal represention of the environment). As well, the role of the decision process of the agent is to select an action (or many of them) accordingly to its properties and to its \protect{\it{intention}}~:~an agent can ``decide'' to climb because he has got the \protect{\it{intention}} to go higher. The notion of \protect{\it{intention}} is here considered as a willing over a property, a complete analysis of this notion, as described in \cite{23},has to be considered as a guide to enforce coherence during the implementation of the final framework. This led us to the definition of the kind of symbol methods have to produce~:~every method should export its influence over agent's properties. So, intentionality in the intentional agent's model is seen as the willing to modify some of its properties accordingly to specific \protect{\it{tendencies}}.\\

Four \protect{\it{tendencies}} have been chosen~:~\protect{\it{increase}},\protect{\it{reduce}},\protect{\it{maintain}} and \protect{\it{independent}}. For example the method \protect{\it{eat}} could be declared~:~\protect{\it{eat()}} \protect{\bf{increase}} \protect{\it{energy}}. This information about the \protect{\it{meaning}} of the method \protect{\it{eat}} can be used by the decision process to select the correct method. There's no need to define at design time which use case could lead the agent to will to increase its energy~:~this is the role of the designer to declare, statically or dynamically, the behavioral rules producing this intention.\\
A simple symbolical and qualitative solver based on those four operators provides a generic decision process~:~whatever properties and action are, they all are treated with this system, for every agent. This allows the selection of one or many action which best respond to agent's intention. There's no guarantee this selection will satisfy every expectation, but such a selection can be done even if it doesn't fit exactly the considered intention. This permits an action selection within an over--constrained context~:~we don't expect an agent to find an exact matching between its intention and its methods. As the environment is highly dynamic, the agent must be able to select a method even if there's no exact solution at this time. We don't want to create an animation from an exact preprocessed solution~:~the simulation is created by the dynamic solving by the agent. Some solution can occur or disappear dymically with the modification of the environment. In this context, trying to always find an exact and complete solution in advance trough a planification process is nonsense.\cite{24}\\
     
Figure \ref{fig3} below shows the architecture of the intentional agent model~:~actions are defined by imperative methods, decision is designed as a set of behavioural rules expressed as logic rules and a link between those two parts is ensured by a qualitative and symbolic solver. This is a first attempt to implement a model based on the framework exposed in section 2~:~the lacking implicit link of fig. 2 is this solver; its semantic towards the model is brought by the notion of intentionality.\\

\begin{figure}[!th]
 \begin{center}
\centerline{\psfig{file=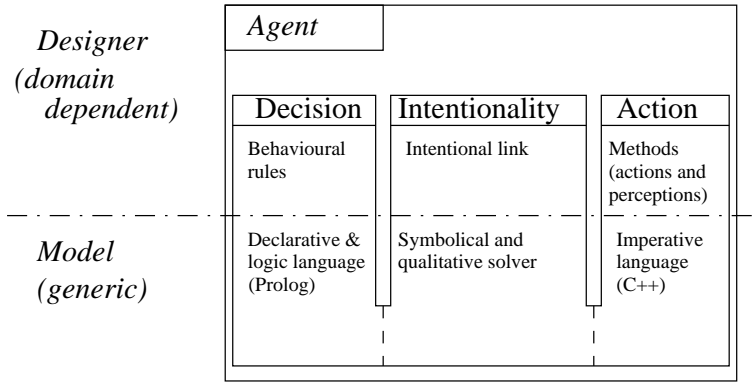}}
\vspace*{8pt}
 \caption[fig3]{\label{fig3}Intentional agent model architecture}
 \end{center}
\end{figure}
    
Let's take the example of a very simple cat's behaviour. Let the method \protect{\it{lookAround}} provides the symbol \protect{\it{danger}}. This method is in charge of the treatment of any state of the cat's environment to provide this symbol if applicable. A dog, a car or anything else dangerous should be correctly perceived as a danger. The symbolical adaptations from the environment to the properties of the agent are processed in the methods.\\
Let some behavioural rules be, asking the cat to eat in order to increase its fitness, and refraining it from eating when it's in danger. This will result in the intention to \protect{\it{reduce}} the \protect{\it{danger}}, if it exists. This intention is not sufficient to select the method \protect{\it{run()}} if there's no further information on its effect on agent's properties, \protect{\it{danger}} in particular.\\
If an information is added to provide the link between action and decision, the method can be selected. So the designer has to define~:~\protect{\it{run()}} \protect{\bf{reduce}} \protect{\it{danger}}.\\
This indirection may seem useless, but it adds some semantical information about the \protect{\it{meaning}} of the action. This is the major benefit of the model~:~this information is specified with a high level of abstraction, using a domain dependent vocabulary.\\
In the same example, we can imagine our cat knows \protect{\it{how}} to mew, but it ``thinks'' it only \protect{\it{increases}} its sex appeal. If at run time, the designer, or any learning process, adds the following information~:~\protect{\it{mew()}} \protect{\bf{reduce}} \protect{\it{danger}}, the corresponding method is now selectable, and an alternative behaviour has been defined. The in--line prototyping through high level and domain dependent concept manipulation has been reached.\\
The information about the symbol provided by the perceptive methods can be used to trigger perception on demand, in order to optimize the whole execution of the behaviour~:~some perception methods are expensive to process but not necessary all the time.\\

Figure \ref{fig4} below sums up the cat's behaviour specification using the intentional model. Keywords \protect{\it{ensure}} and \protect{\it{provide}} are used to specify links between respectively perception and decision in the first place, and decision and action in the second place.

\begin{figure}[!th]
\begin{center}
\begin{tabular}{|l|l|l|}
\hline
\bf{Declarative}        & Behavioural             & \tt{main :- eat.}  \\                      
                        &  rules                  & \tt{eat :- not(danger).}\\ 
\hline          
\bf{Intentional}       & Perception/decision      & \tt{lookAround provide : danger} \\           
                       & link                     &\\
\cline {2-3}         
                       & Action/decision          & \tt{run ensure : reduce danger}   \\
                       & link                     & \tt{mew ensure : increase sexAppeal,}\\
                       &                          & \tt{~~~~~~~~~~~~~reduce danger}  \\
\hline	             
\bf{Imperative}        & Perception               & \tt{void lookAround(void) \{...\} }            \\
\cline{2-3}          
                       & Action                   & \tt{void run(void) \{...\}}  \\              
                       &                          & \tt{void mew(void) \{...\}}  \\             
\hline
\end{tabular}
\caption[fig4]{\label{fig4}Simple cat's behaviour specification sum up (pseudo code)}
\end{center}
\end{figure}

\section{Implementation}
The first step of the application of such a proposition is the building of a tool able to execute it. Primary tests have been led using \protect{\tt{oRis}} and \protect{\tt{Prolog}}. \protect{\tt{oRis}} is an imperative and interpreted language. It provides many agent oriented functions~:~scheduler for parallel activities, message exchange protocol between entities, dynamic code modification,\ldots This brings most of the needed functions of our model. We wrapped \protect{\tt{GNU Prolog}}'s engine in \protect{\tt{oRis}} to add the required first logic order capabilities. The wrapper offers the ability to create many prolog context~:~every agent owns a separate set of stacks to handle all fact and predicate it needs. When a query is performed, the appropriate context is restored within the prolog engine. This is totally transparent to the user which can consider that every agent owns a separated prolog engine. The main asset of this implementation is that there's only one instance of prolog, statically linked with the application~:~performance remains correct when many agents are implied in the simulation. As this is an add--on to \protect{\tt{oRis}}, it doesn't provide all expressiveness we expect~:~those two languages weren't designed with the idea of such an interface. Another problem is that \protect{\tt{GNU Prolog}} is monolithic~:~when starting a query, you have to wait for the first result; it's hard to schedule many agents as their behaviour become more complex. This first implementation was only a test for a preliminary try in order to validate the basis of our approach.\\
We are now working with an implementation based on our new \protect{\tt{C++}} library, \protect{\tt{AR\'eVi}}, which offers scheduling and rendering features together with \protect{\tt{SWI Prolog}} which is capable of multi--threading resolution. This allows us to explore deeper the proposition of the intentional model. But it is still a test plateform~:~this implementation only provide a simple bridge between \protect{\tt{C++}} and \protect{\tt{Prolog}}, \protect{\it{e.g.~:}}~the predicate \protect{\it{getProperty(+PropertyName, ?Value)}} has been created to get at any time the value of a \protect{\tt{C++}} property from \protect{\tt{Prolog}} part. Some other simple mechanisms allow to give \protect{\it{tendencies}} back to the \protect{\tt{C++}} simulation engine from the \protect{\tt{Prolog}} decision process. If a first implementation of intentional agent model is possible with the help of this tool, there is a lack of expressivness of the so created framework because of the deep difference between the two involved language. The binding between \protect{\tt{C++}} attribute and \protect{\tt{Prolog}} variable is still partially handled by the designer.\\ 

In parallel, we now are building a new language from scratch based on \protect{\tt{oRis}} and \protect{\tt{AR\'eVI}} including in its virtual machine the Warren's abstract machine instruction set.\cite{25} This will allow a native support of the needed link between the two aspects of the behavioural specification, with a deep control of the scheduling of mixed imperative / declarative code. This tool will be able to execute any kind of model based on our mixed approach of behaviour specification.\\

\section{Conclusion}
We have presented our reflexion around the need of a greater coherence in the behaviour specification process. This increases the coherence of the whole framework and enables a high level in--line prototyping with a reduced set of concepts~:~the user interaction with the behavioural model is the same at design time and at run--time. This is one more step toward the interactive prototyping of an agent's behaviour by a non computer science expert. This need should be considered from the early begining of the building  of an execution plateform in order to provide an easy interface to the designer. This approach led us to the proposition of a new tool and to the specification of the model we could simulate within this framework~:~intentional agent model. The building of a new tool from scratch ensures the usage of the same concepts from conception to execution. We consider it is the only way to reach the expressivness we are targeting at.\\ 

Another direction of our further work is the behavioural inheritance~:~as the behaviour is totally defined by a set of logic rules, it's easier to modify or enrich. Therefore, we now want to define some inheritance mechanisms to allow an agent to inherit --- statically or dynamicly -- and to share its knowledge. For a pr\-ey~/~pre\-da\-tor example, it would permit the creation of a new kind of predator at run--time of the behaviour of a standard predator, and the addition or deletion of some clauses of the \protect{\it{purchase}} rule.\\

The last direction we'd like to explore is a direct benefit of the semantic added in the model. As the \protect{\it{meaning}} of methods is partially expressed within the model, we could expect a greater easiness in the application of symbolical learning and explanation generation process with intentional agent.


%bibliographystyle{spiebib}
%bibliography{article}

\section*{Photo and Bibliography}

\vspace*{13pt}
\noindent
\parbox{5truein}{
\begin{minipage}[b]{1truein}
\centerline{{\psfig{file=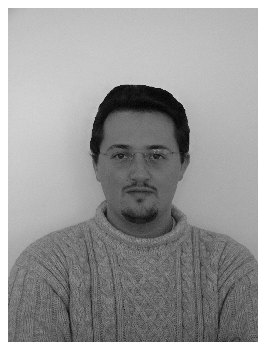}}}
\end{minipage} 
\hfill         %to 2nd column
\begin{minipage}[b]{3.85truein}
{{\bf Pierre-Alexandre FAVIER} is PhD. student in computer science
at the ENIB, Brest, France. He received his diploma in computer science 
engineering from ENIB in 2001. He works in the Plateform for Virtual
Reality project of the CERV.

\hglue 15pt His research interests and teaching activities are in the areas
of artificial intelligence and virtual reality.}
\end{minipage} } %close for parbox

\vspace*{13pt}	%SECOND PHOTO
\noindent
\parbox{5truein}{
\begin{minipage}[b]{1truein}
\centerline{{\psfig{file=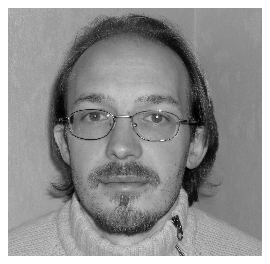}}}
\end{minipage} 
\hfill         %to 2nd column
\begin{minipage}[b]{3.85truein}
{{\bf Pierre DE LOOR} is Senior Lecturer at the ENIB from 1998. 
He received the PhD degrees in cybernetic and computer science 
from the University of Reims, France, in 1996. He works in the 
Interactive and Participative Simulation project of the CERV. 

\hglue 15pt His research and teaching interests include artificial intelligence, 
automatic learning, human factor simulation, and distributed 
heuristics.}
\end{minipage} } %close for parbox 
\vfill\eject

\end{document}